\documentclass[runningheads]{llncs}
\usepackage{graphicx}
\usepackage{diagbox} 
\usepackage{cite} 
\usepackage[table,xcdraw]{xcolor} 
\usepackage{subcaption} 
\usepackage{multirow} 
\usepackage{comment} 

\usepackage{tikz}
\usepackage[absolute,overlay]{textpos}
\usepackage{blindtext}
\usepackage{relsize}
\usepackage{hyperref}

\tikzset{fontscale/.style={font=\relsize{#1}}}

\def\BibTeX{{\rm B\kern-.05em{\sc i\kern-.025em b}\kern-.08em
    T\kern-.1667em\lower.7ex\hbox{E}\kern-.125emX}}

\begin{document}
\title{Exploring fully convolutional networks for the segmentation of hyperspectral imaging applied to advanced driver assistance systems\thanks{This work was partially supported by the Basque Government under grants  PIBA-2018-1-0054, KK-2021/00111 and PRE 2021 1 0113 and by the Spanish Ministry of Science and Innovation under grant PID2020-115375RB-I00. We thank the University of the Basque Country for allocation of computational resources.}}

\titlerunning{Exploring FCNs for the segmentation of HSI applied to ADAS}
%
\author{Jon Gutiérrez-Zaballa\inst{1}\orcidID{0000-0002-6633-4148} \and \\ Koldo Basterretxea\inst{2}\orcidID{0000-0002-5934-4735} \and \\ Javier Echanobe\inst{3}\orcidID{0000-0002-1064-2555} \and \\ M. Victoria Martínez\inst{3} \and \\ Inés del Campo \inst{3}\orcidID{0000-0002-6378-5357}}
\authorrunning{J. Gutiérrez-Zaballa et al.}
%
\institute{Jon Gutiérrez-Zaballa is with the Department of Electronics Technology, \\ University of the Basque Country, 48013, Bilbao, Spain \email{j.gutierrez@ehu.eus} \and Researcher is with the Department of Electronics Technology, \\ University of the Basque Country, 48013, Bilbao, Spain \and Researchers are with the Department of Electricity and Electronics, \\
University of the Basque Country, 48940 Leioa, Spain}

\maketitle              

\begin{textblock*}{21cm}(1.5cm,26cm)
\begin{tikzpicture}
    \draw (0,0) rectangle (18,0.5); 
    \end{tikzpicture}
\end{textblock*} 

\begin{textblock*}{21cm}(0cm,26cm)
\begin{tikzpicture}
    \node (center) {c};
    \path (center)+(10.5,4) node [fontscale=-1] (name) {\copyright 2024 Springer Nature. Final published version of the article can be found at \href{https://link.springer.com/chapter/10.1007/978-3-031-12748-9\_11}{10.1007/978-3-031-12748-9\_11}.};
    \end{tikzpicture}
\end{textblock*} 

\begin{abstract}
Advanced Driver Assistance Systems (ADAS) are designed with the main purpose of increasing the safety and comfort of vehicle occupants. Most of current computer vision-based ADAS perform detection and tracking tasks quite successfully under regular conditions, but are not completely reliable, particularly under adverse weather and changing lighting conditions, neither in complex situations with many overlapping objects. In this work we explore the use of hyperspectral imaging (HSI) in ADAS on the assumption that the distinct near infrared (NIR) spectral reflectances of different materials can help to better separate the objects in a driving scene. In particular, this paper describes some experimental results of the application of fully convolutional networks (FCN) to the image segmentation of HSI for ADAS applications. More specifically, our aim is to investigate to what extent the spatial features codified by convolutional filters can be helpful to improve the performance of HSI segmentation systems. With that aim, we use the HSI-Drive v1.1 dataset, which provides a set of labelled images recorded in real driving conditions with a small-size snapshot NIR-HSI camera. Finally, we analyze the implementability of such a HSI segmentation system by prototyping the developed FCN model together with the necessary hyperspectral cube preprocessing stage and characterizing its performance on an MPSoC.

\end{abstract}

\section{Introduction}
Today, thanks to the availability of small-size, portable, snapshot hyperspectral cameras, it is possible to set-up HSI processing systems on moving platforms. The use of drones for precision agriculture and ecosystem monitoring is probably one of the most active and mature application domains \cite{govender2007review}. The research into how hyperspectral information can be used to develop more capable and robust ADAS is, on the contrary, in its infancy \cite{huang2020hyperspectral, winkens2017hyko, winkens2017hyperspectral}. HSI provides rich information about how materials reflect light of different wavelengths (spectral reflection), and this can be used to identify and classify surfaces and objects in an scene. Thus, with the application of appropriate information processing techniques, HSI can help to enhance the accuracy and robustness of current ADAS for object identification and tracking and, eventually, can be used for scene understanding, which is a step forward in the achievement of more capable and intelligent ADS (Autonomous Driving Systems).

HSI segmentation of real driving scenes is, however, challenging for a variety of reasons. First, the spectral reflectance signatures of the different objects, e.g. metallic white vehicles bodies and road marks, may be weakly separable. Second, extracting spatial features that could help segmenting items with similar spectral reflectances is difficult as a consequence of the enormous diversity of shapes, view angles and scales. Finally, it should be always kept in mind that developed segmentation algorithms need to be computed with very demanding latency requirements on resource constrained onboard processing platforms.

In this article we describe some results of a research that investigates how FCN can be applied to enhance the segmentation accuracy of images acquired in real driving scenarios with a small-size mosaic snapshot hyperspectral camera. We present a simple application example of scene understanding for the separation of the drivable (tarmac) and non-drivable areas (identifying sky and vegetation) in the acquired image sequences as well as for the recognition of road marks, which could be used to enhance automatic lane keeping and trajectory planning systems for ADS. Finally, we describe the rapid prototyping workflow used to develop a functional HSI segmentation processing system on a Xilinx Zynq UltraScale MPSoC, from algorithm exploration and model optimization to the final implementation.

\section{Experimental Setup}\label{sec:expSetup}
When dealing with a semantic segmentation problem, it is of utmost importance to adapt the structure of the neural network to the unique characteristics of the dataset. Thus, once the suitability of the dataset has been verified, a hyperparameter tuning and optimization process should be carried out on the neural network.

\subsection{The Dataset}
As it is reported in \cite{basterretxeahsi}, there are very few datasets of hyperspectral imagery for ADAS and ADS applications, one of which is precisely presented in \cite{basterretxeahsi}, HSI Drive. HSI Drive v1.1 contains 276 images of urban, road and highway scenarios in diverse weather (sunny, cloudy, rainy and foggy) and lightning (dawn, midday, sunset) conditions taken during Spring (121 images) and Summer (155 images).

The driving scenes have been recorded with a Photonfocus camera that includes an Imec 25-band VIS-NIR (535nm-975nm) sensor based on a CMOSIS CMV200 image wafer sensor. The global resolution is 1088 x 2048 pixels with 5$\mu m$ x 5$\mu m$ size. However, as the spectral bands are extracted from a mosaic formed by 5x5 pixel window Fabri-Perot filters, the final resolution of the HSI cubes is 216 x 409 x 25 \cite{mv1-d2048x1088-hs02-96-g2}. This implies including a preprocessing stage in the processing pipeline that is addressed in Subsection \ref{subsec:imagePre}.

The original labelling separates the scenes into 10 classes taking into account the surface reflectances of the materials. Those classes are: Road, Road Marks, Vegetation, Painted Metal, Sky, Concrete/Stone/Brick, Pedestrian/Cyclist, Water, Unpainted Metal and Glass/Transparent Plastic. Furthermore, it has to be noted that the labelling of the dataset has followed a weak approach in order to provide the network with the most precise data. This means, for example, that pixels that are in the junction of two or more surfaces have been left out of the labelling process. However, these pixels do take part in the training process of a convolutional network. In fact, the training of convolutional neural networks with weakly labelled datasets is a line of research itself \cite{wang2020weakly}.

Spectral separability analysis measures the differences in the surface reflectance patterns of the materials that belong to different classes, which is an index of how well a semantic classifier could perform. One of the most common criteria in remote-sensing applications is JeffreysMatusita distance \cite{forestier2013comparison}. It ranges from 0 to 2 but does not have a linear interpretation as 0-1 values mean very poor separability, 1.0-1.9 values account for moderate separability and 1.9-2.0 values indicate good separability \cite{basterretxeahsi}.

\begin{table}[h!]
\centering
\caption{JeffreysMatusita (JM) interclass distances.}
\label{tab:separability}
\resizebox{10.25cm}{!}{%
\begin{tabular}{c|c|c|c|c|c|c|c|c|c|}
\cline{2-10}
 & \textbf{Road} & \textbf{Road M.} & \textbf{Veg.} & \textbf{P. Met.} & \textbf{Sky} & \textbf{Conc.} & \textbf{Ped.} & \textbf{Unp. Met.} & \textbf{Glass} \\ \hline
\multicolumn{1}{|c|}{\textbf{Road}} &  & 1.92 & 1.83 & 1.65 & 1.98 & 1.44 & 1.84 & 1.42 & 1.49 \\ \cline{1-1}
\multicolumn{1}{|c|}{\textbf{Road Marks}} & 1.92 &  & 1.79 & 1.63 & 1.87 & 1.68 & 1.92 & 1.92 & 1.93 \\ \cline{1-1}
\multicolumn{1}{|c|}{\textbf{Vegetation}} & 1.83 & 1.79 &  & 1.44 & 1.96 & 1.64 & 1.81 & 1.73 & 1.81 \\ \cline{1-1}
\multicolumn{1}{|c|}{\textbf{Painted Metal}} & 1.65 & 1.63 & 1.44 &  & 1.91 & 1.48 & 1.70 & 1.35 & 1.48 \\ \cline{1-1}
\multicolumn{1}{|c|}{\textbf{Sky}} & 1.98 & 1.87 & 1.96 & 1.91 &  & 1.97 & 1.98 & 1.98 & 1.80 \\ \cline{1-1}
\multicolumn{1}{|c|}{\textbf{Concrete}} & 1.44 & 1.68 & 1.64 & 1.48 & 1.97 &  & 1.74 & 1.62 & 1.68 \\ \cline{1-1}
\multicolumn{1}{|c|}{\textbf{Pedestrian}} & 1.84 & 1.92 & 1.81 & 1.70 & 1.98 & 1.74 &  & 1.79 & 1.66 \\ \cline{1-1}
\multicolumn{1}{|c|}{\textbf{Unpainted Metal}} & 1.42 & 1.92 & 1.73 & 1.35 & 1.98 & 1.62 & 1.79 &  & 1.38 \\ \cline{1-1}
\multicolumn{1}{|c|}{\textbf{Glass}} & 1.49 & 1.93 & 1.81 & 1.48 & 1.80 & 1.68 & 1.66 & 1.38 &  \\ \hline
\multicolumn{1}{|c|}{\textbf{Mean}} & \textbf{1.73} & \textbf{1.80} & \textbf{1.78} & \textbf{1.63} & \textbf{1.96} & \textbf{1.69} & \textbf{1.82} & \textbf{1.68} & \textbf{1.71} \\ \hline
\end{tabular}}
\end{table}

Table \ref{tab:separability} sums up the JM interclass distances of the ten classes. As it can be observed, the separability of some of the classes such as Road/Road Marks (1.92), Road/Sky (1.98) or Road Marks/Unpainted Metal (1.92) is promisingly high while classes like Road/Concrete (1.44), Road Marks/Painted Metal (1.63) and Painted Metal/Unpainted Metal (1.35) show low separability indexes.

\subsection{FCNs for HSI Image Segmentation}
The neural network selected to perform semantic segmentation is a typical FCN known as U-Net \cite{unet} which was originally intended for biological image segmentation but has been widely used for other segmentation tasks, such as, precision agriculture \cite{wang2020weakly} and aerial city recognition \cite{cui2019multiscale}. The idea of using a FCN is to combine the intrinsic spectral characteristics of the different classes with the spatial relationships that should be extracted by the convolution operations. 

We have adapted the original architecture of the U-Net \cite{unet} to the unique characteristics of the dataset to achieve the best trade-off between segmentation performance and computational complexity. With this aim we have performed a grid search of the optimum combination of model hyperparameters by evaluating the segmentation accuracy on a subset of 45 images selected from all possible environment/weather conditions. 

\begin{figure}[h!]
\centering
\includegraphics[height=5.85cm]{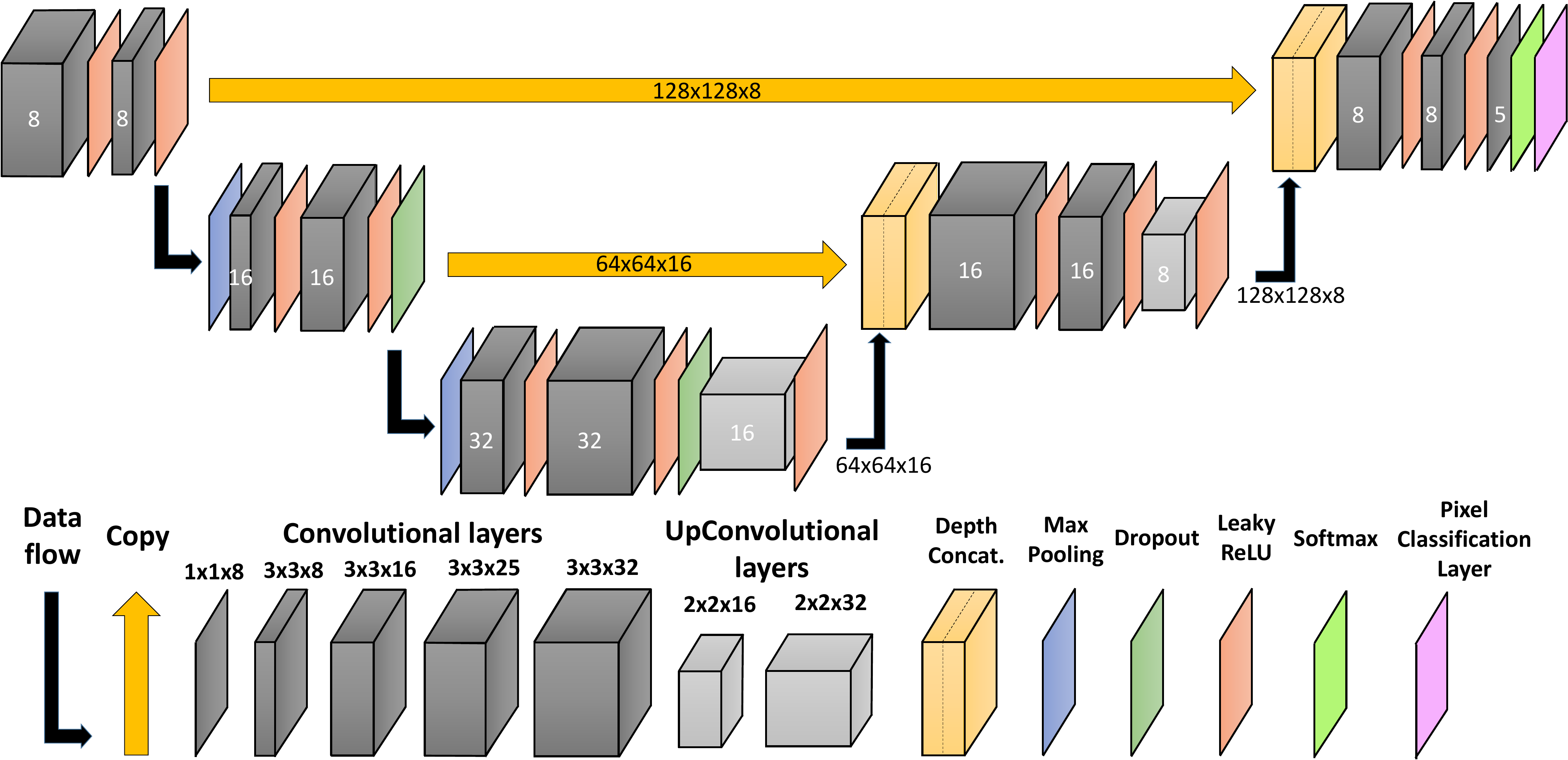}
\caption{Architecture of the modified U-Net.} \label{fig:miUnet}
\end{figure}

The set of analyzed hyperparameters included: the size of the input image patches, the overlapping between patches, the encoder depth and the number of filters in the first convolutional block. In order to avoid an unaffordable optimization time, we have consulted the typical values of the hyperparameters to be optimized in the literature \cite{wang2020weakly}. This way, a specific range has been set for each hyperparameter. Specifically, the value of the encoder depth has been varied between 2 and 4, the number of initial filters between 8 and 32 (in powers of 2), while for the patch size the values 64 and 128 have been evaluated.

The values of the hyperparameters that have output the three best results are: 2-3-4 for the encoder/decoder depth, 8-16-16 for the initial number of filters and 128-128-128 for the side of the square patch. According to the accuracy/complexity trade-off criterion we have selected the 2/8/128 set (Figure \ref{fig:miUnet} shows the final architecture of the network). As a consequence of the size of the patch and to benefit from the effect of overlapping, it has been decided to divide the input test images in 18 (3 x 6) patches. 

\section{Segmentation Results}\label{sec:segRes}
The first experiment focuses on segmenting 3 classes: Road, Road Marks and No Drivable (the remaining classes). The proposed low-complexity segmentation system would be aimed at a possible final system for the discrimination of drivable and non-drivable zones, together with a lane-keeping aid.

In a second experiment we have added two additional classes to the model training; Vegetation and Sky. These two categories have been selected due to their satisfactory spectral separability indexes (see Table \ref{tab:separability}). The exploration of more complex segmentation models including all classes in the dataset has also been performed but obtained results are irregular and not concluding, and will require further investigation.

In order to perform a neural network training over this dataset, the 276 images have to be divided into training, validation and test subsets. This division has been performed as follows: 162 images for training, 57 for validation and 57 for testing, preserving class proportionality in all the three subsets.

The chosen metrics to evaluate the segmentation ability of the neural network are accuracy, precision and intersection over union (IoU). As Equations \ref{equ:accuracy}, \ref{equ:precision} and \ref{equ:iou} show, accuracy accounts for the false negatives (FN), precision takes into account the false positives (FP) and IoU combines both aspects:

\begin{equation}
  A_i = \frac{TP_i}{TP_i + FN_{i}}
  \label{equ:accuracy}
\end{equation}

\begin{equation}
  P_i = \frac{TP_i}{TP_i + FP_{i}}
  \label{equ:precision}
\end{equation}

\begin{equation}
IoU_i = \frac{TP_i}{TP_i + FN_{i} + FP_{i}}
\label{equ:iou}
\end{equation}

where $i$ is the class index such that, for example, $FN_{i}$ accounts for the pixels that have been predicted as not belonging to class i, but are actually part of class i. \\

As a consequence of the dataset being heavily imbalanced (the number of pixels in the test dataset is: Road 2,067,379; Road M. 99,426; Veget. 820,804; Sky 163,127 and Other 363,345) it is useful not to only represent the global metrics but also the mean values and, more specifically, the weighted scores. In order to do that, some weighting factors, which are related to the inverse of the frequency of the classes in the dataset, have been previously computed.

\subsection{U-Net}
Table \ref{tab:metricsFloatingPoint} collects the performance, in accordance with the above mentioned metrics, of the modified U-Net and also the segmentation metrics after the overlapped patches have been joined to reconstruct the images to their original resolution. The comparison depicts that the use of overlapping patches improves the segmentation, specially the precision, compared to the case in which the patches do not overlap. This is because neural networks tend to fail to predict the pixels of the patch contours because they lack surrounding information.

\begin{table}[h!]
\centering
\caption{Performance of the modified U-Net (patches and overlapping patches) and ANN on the 3-classes (up) and 5-classes (down) test datasets.}
\label{tab:metricsFloatingPoint}
\resizebox{11cm}{!}{%
\begin{tabular}{cccccccccc}
\cline{2-10}
\multicolumn{1}{c|}{} & \multicolumn{6}{c|}{\textbf{U-Net}} & \multicolumn{3}{c|}{\textbf{ANN}} \\ \cline{2-10} 
\multicolumn{1}{c|}{} & \multicolumn{3}{c|}{\textbf{\begin{tabular}[c]{@{}c@{}}Patches\\ (128x128x25)\end{tabular}}} & \multicolumn{3}{c|}{\textbf{\begin{tabular}[c]{@{}c@{}}Rebuilt images from\\ overlapping patches\end{tabular}}} & \multicolumn{3}{c|}{\textbf{\begin{tabular}[c]{@{}c@{}}Pixels\\ (1x1x25)\end{tabular}}} \\ \cline{2-10} 
\multicolumn{1}{c|}{} & \multicolumn{1}{c|}{\textbf{Accuracy}} & \multicolumn{1}{c|}{\textbf{Precision}} & \multicolumn{1}{c|}{\textbf{IoU}} & \multicolumn{1}{c|}{\textbf{Accuracy}} & \multicolumn{1}{c|}{\textbf{Precision}} & \multicolumn{1}{c|}{\textbf{IoU}} & \multicolumn{1}{c|}{\textbf{Accuracy}} & \multicolumn{1}{c|}{\textbf{Precision}} & \multicolumn{1}{c|}{\textbf{IoU}} \\ \hline
\multicolumn{1}{|c|}{\textbf{Road}} & \multicolumn{1}{c|}{97.90} & \multicolumn{1}{c|}{95.66} & \multicolumn{1}{c|}{93.74} & \multicolumn{1}{c|}{98.54} & \multicolumn{1}{c|}{94.56} & \multicolumn{1}{c|}{93.25} & \multicolumn{1}{c|}{85.10} & \multicolumn{1}{c|}{92.51} & \multicolumn{1}{c|}{79.62} \\ \hline
\multicolumn{1}{|c|}{\textbf{Road Marks}} & \multicolumn{1}{c|}{90.25} & \multicolumn{1}{c|}{73.11} & \multicolumn{1}{c|}{67.75} & \multicolumn{1}{c|}{87.89} & \multicolumn{1}{c|}{77.22} & \multicolumn{1}{c|}{69.80} & \multicolumn{1}{c|}{68.10} & \multicolumn{1}{c|}{21.90} & \multicolumn{1}{c|}{19.86} \\ \hline
\multicolumn{1}{|c|}{\textbf{No Drivable}} & \multicolumn{1}{c|}{91.07} & \multicolumn{1}{c|}{97.16} & \multicolumn{1}{c|}{88.71} & \multicolumn{1}{c|}{91.20} & \multicolumn{1}{c|}{98.57} & \multicolumn{1}{c|}{90.01} & \multicolumn{1}{c|}{86.46} & \multicolumn{1}{c|}{89.32} & \multicolumn{1}{c|}{78.42} \\ \hline
\multicolumn{1}{|c|}{\textbf{Overall}} & \multicolumn{1}{c|}{95.37} & \multicolumn{1}{c|}{95.55} & \multicolumn{1}{c|}{91.31} & \multicolumn{1}{c|}{95.42} & \multicolumn{1}{c|}{95.44} & \multicolumn{1}{c|}{91.50} & \multicolumn{1}{c|}{85.14} & \multicolumn{1}{c|}{89.32} & \multicolumn{1}{c|}{77.46} \\ \hline
\multicolumn{1}{|c|}{\textbf{Mean}} & \multicolumn{1}{c|}{93.07} & \multicolumn{1}{c|}{88.64} & \multicolumn{1}{c|}{83.40} & \multicolumn{1}{c|}{92.54} & \multicolumn{1}{c|}{90.12} & \multicolumn{1}{c|}{84.35} & \multicolumn{1}{c|}{79.89} & \multicolumn{1}{c|}{67.93} & \multicolumn{1}{c|}{59.30} \\ \hline
\multicolumn{1}{|c|}{\textbf{Weighted}} & \multicolumn{1}{c|}{90.60} & \multicolumn{1}{c|}{75.71} & \multicolumn{1}{c|}{70.27} & \multicolumn{1}{c|}{88.54} & \multicolumn{1}{c|}{79.43} & \multicolumn{1}{c|}{72.60} & \multicolumn{1}{c|}{70.04} & \multicolumn{1}{c|}{29.36} & \multicolumn{1}{c|}{26.27} \\ \hline
 &  &  &  &  &  &  &  &  &  \\ \hline
\multicolumn{1}{|c|}{\textbf{Road}} & \multicolumn{1}{c|}{92.61} & \multicolumn{1}{c|}{99.05} & \multicolumn{1}{c|}{91.36} & \multicolumn{1}{c|}{93.28} & \multicolumn{1}{c|}{99.00} & \multicolumn{1}{c|}{92.41} & \multicolumn{1}{c|}{73.29} & \multicolumn{1}{c|}{94.30} & \multicolumn{1}{c|}{70.18} \\ \hline
\multicolumn{1}{|c|}{\textbf{Road Marks}} & \multicolumn{1}{c|}{80.93} & \multicolumn{1}{c|}{75.39} & \multicolumn{1}{c|}{64.02} & \multicolumn{1}{c|}{78.32} & \multicolumn{1}{c|}{79.11} & \multicolumn{1}{c|}{64.90} & \multicolumn{1}{c|}{68.74} & \multicolumn{1}{c|}{17.50} & \multicolumn{1}{c|}{16.21} \\ \hline
\multicolumn{1}{|c|}{\textbf{Vegetation}} & \multicolumn{1}{c|}{94.98} & \multicolumn{1}{c|}{94.63} & \multicolumn{1}{c|}{90.12} & \multicolumn{1}{c|}{95.74} & \multicolumn{1}{c|}{95.80} & \multicolumn{1}{c|}{91.88} & \multicolumn{1}{c|}{93.84} & \multicolumn{1}{c|}{91.47} & \multicolumn{1}{c|}{86.29} \\ \hline
\multicolumn{1}{|c|}{\textbf{Sky}} & \multicolumn{1}{c|}{97.86} & \multicolumn{1}{c|}{93.09} & \multicolumn{1}{c|}{91.23} & \multicolumn{1}{c|}{97.49} & \multicolumn{1}{c|}{93.39} & \multicolumn{1}{c|}{91.20} & \multicolumn{1}{c|}{91.04} & \multicolumn{1}{c|}{74.53} & \multicolumn{1}{c|}{69.44} \\ \hline
\multicolumn{1}{|c|}{\textbf{Other}} & \multicolumn{1}{c|}{84.97} & \multicolumn{1}{c|}{62.71} & \multicolumn{1}{c|}{56.45} & \multicolumn{1}{c|}{84.83} & \multicolumn{1}{c|}{64.59} & \multicolumn{1}{c|}{57.90} & \multicolumn{1}{c|}{56.18} & \multicolumn{1}{c|}{42.94} & \multicolumn{1}{c|}{32.17} \\ \hline
\multicolumn{1}{|c|}{\textbf{Overall}} & \multicolumn{1}{c|}{91.79} & \multicolumn{1}{c|}{93.29} & \multicolumn{1}{c|}{86.47} & \multicolumn{1}{c|}{92.75} & \multicolumn{1}{c|}{93.80} & \multicolumn{1}{c|}{87.66} & \multicolumn{1}{c|}{77.02} & \multicolumn{1}{c|}{85.24} & \multicolumn{1}{c|}{68.45} \\ \hline
\multicolumn{1}{|c|}{\textbf{Mean}} & \multicolumn{1}{c|}{90.18} & \multicolumn{1}{c|}{84.98} & \multicolumn{1}{c|}{78.64} & \multicolumn{1}{c|}{89.93} & \multicolumn{1}{c|}{86.38} & \multicolumn{1}{c|}{79.66} & \multicolumn{1}{c|}{76.62} & \multicolumn{1}{c|}{64.15} & \multicolumn{1}{c|}{54.86} \\ \hline
\multicolumn{1}{|c|}{\textbf{Weighted}} & \multicolumn{1}{c|}{88.64} & \multicolumn{1}{c|}{82.14} & \multicolumn{1}{c|}{75.27} & \multicolumn{1}{c|}{87.40} & \multicolumn{1}{c|}{84.15} & \multicolumn{1}{c|}{75.93} & \multicolumn{1}{c|}{75.28} & \multicolumn{1}{c|}{44.00} & \multicolumn{1}{c|}{39.55} \\ \hline
\end{tabular}}
\end{table}

\begin{figure}[h!]
\begin{subfigure}{0.32\linewidth}
\centering
\includegraphics[height=1.6cm]{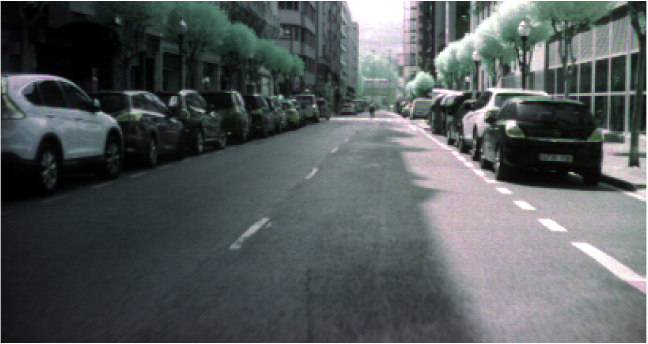}
\caption{Urban, visible.}
\label{fig:visUrban}
\end{subfigure}
\begin{subfigure}{0.32\linewidth}
\centering
\includegraphics[height=1.6cm]{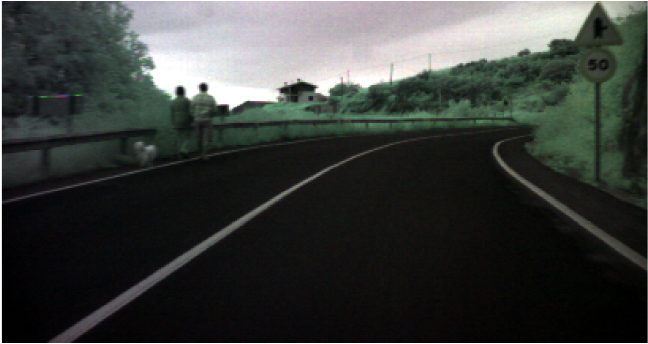}
\caption{Road, visible.}
\label{fig:visRoad}
\end{subfigure}
\begin{subfigure}{0.32\linewidth}
\centering
\includegraphics[height=1.6cm]{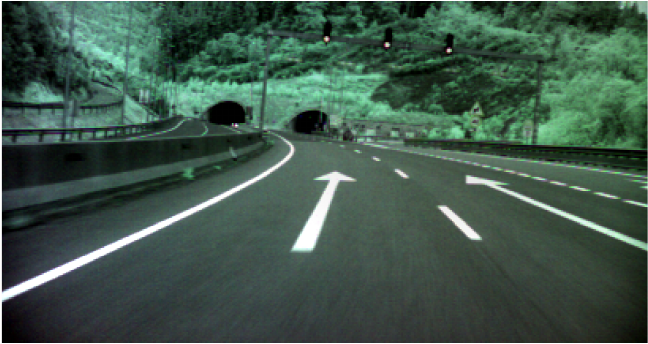}
\caption{Highway, visible.}
\label{fig:visHighway}
\end{subfigure}\\[1ex]
\begin{subfigure}{0.32\linewidth}
\centering
\includegraphics[height=1.9cm]{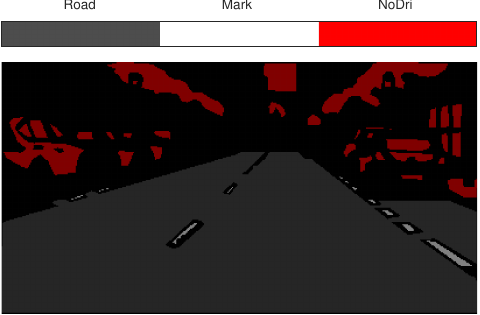}
\caption{Urban, GT (3 class).}
\label{fig:gtUrban3clases}
\end{subfigure}%
\begin{subfigure}{0.32\linewidth}
\centering
\includegraphics[height=1.9cm]{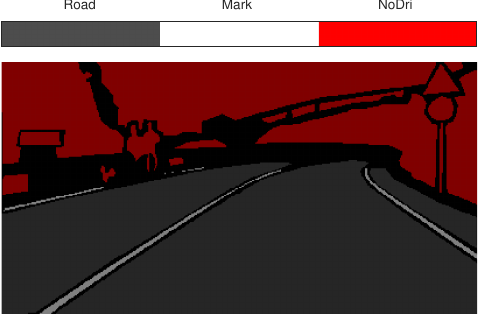}
\caption{Road, GT (3 class).}
\label{fig:gtRoad3clases}
\end{subfigure}
\begin{subfigure}{0.32\linewidth}
\centering
\includegraphics[height=1.9cm]{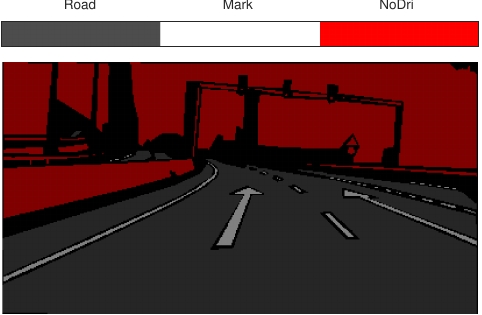}
\caption{Highway, GT (3 class).}
\label{fig:gtHighway3clases}
\end{subfigure}\\[1ex]
\begin{subfigure}{0.32\linewidth}
\centering
\includegraphics[height=1.9cm]{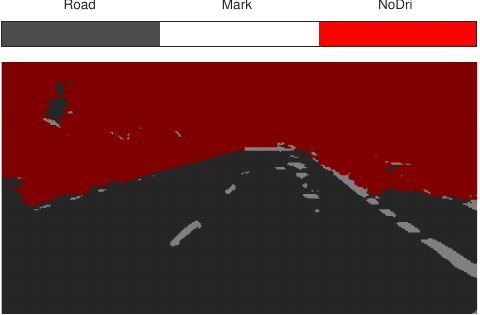}
\caption{Urban, seg. (3 class).}
\label{fig:segUrban3clases}
\end{subfigure}%
\begin{subfigure}{0.32\linewidth}
\centering
\includegraphics[height=1.9cm]{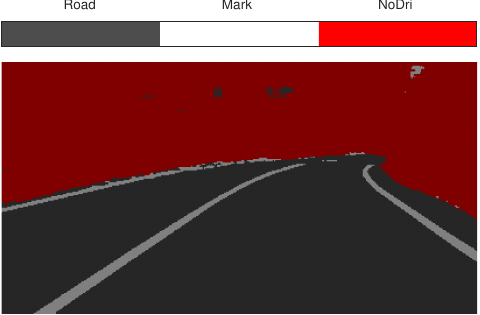}
\caption{Road, seg. (3 class).}
\label{fig:segRoad3clases}
\end{subfigure}
\begin{subfigure}{0.32\linewidth}
\centering
\includegraphics[height=1.9cm]{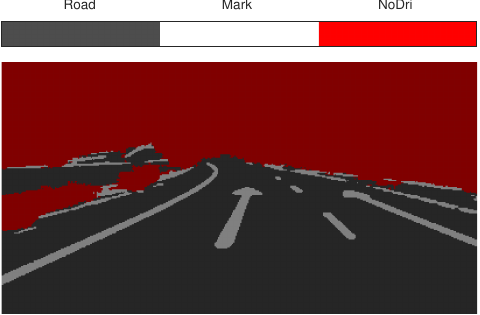}
\caption{Highway, seg. (3 class)}
\label{fig:segHighway3clases}
\end{subfigure}\\[1ex]
\begin{subfigure}{0.32\linewidth}
\centering
\includegraphics[height=1.9cm]{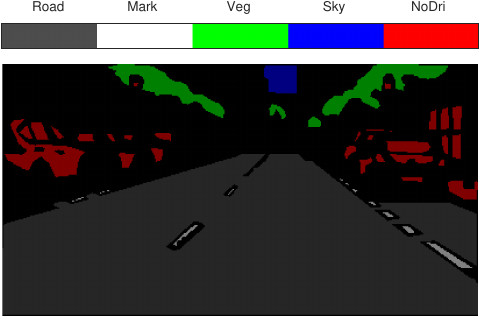}
\caption{Urban, GT (5 class).}
\label{fig:gtUrban5clases}
\end{subfigure}
\begin{subfigure}{0.32\linewidth}
\centering
\includegraphics[height=1.9cm]{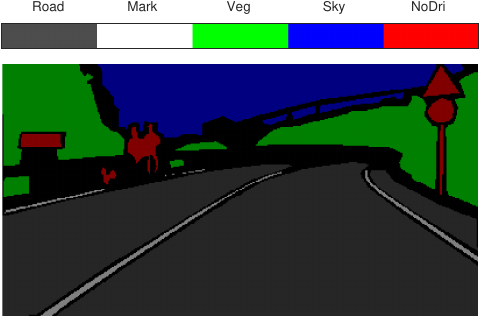}
\caption{Road, GT (5 class).}
\label{fig:gtRoad5clases}
\end{subfigure}
\begin{subfigure}{0.32\linewidth}
\centering
\includegraphics[height=1.9cm]{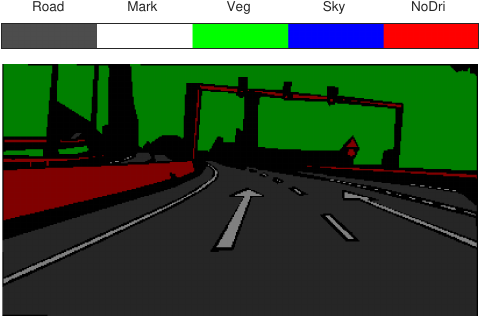}
\caption{Highway, GT (5 class).}
\label{fig:gtHighway5clases}
\end{subfigure}\\[1ex]
\begin{subfigure}{0.32\linewidth}
\centering
\includegraphics[height=1.9cm]{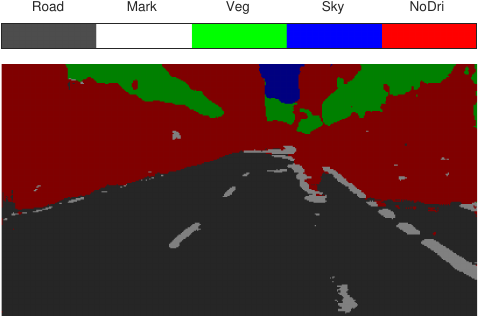}
\caption{Urban, seg. (5 class).}
\label{fig:segUrban5clases}
\end{subfigure}%
\begin{subfigure}{0.32\linewidth}
\centering
\includegraphics[height=1.9cm]{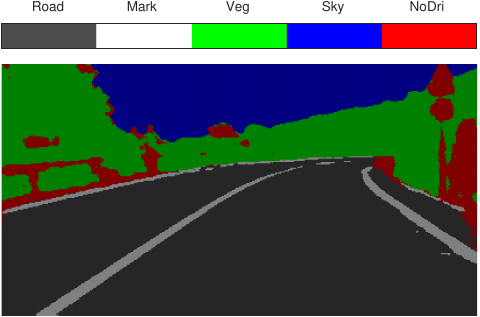}
\caption{Road, seg. (5 class).}
\label{fig:segRoad5clases}
\end{subfigure}
\begin{subfigure}{0.32\linewidth}
\centering
\includegraphics[height=1.9cm]{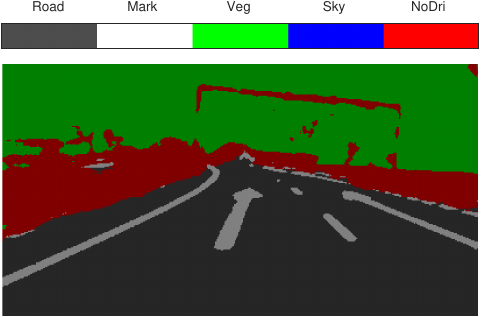}
\caption{Highway, seg. (5 class).}
\label{fig:segHighway5clases}
\end{subfigure}
\caption{Comparison among the visible (first row), 3-class ground truth (second row), 3-class U-Net segmentation (third row), 5-class ground truth (fourth row) and 5-class U-Net segmentation (fifth row) images of three different scenarios: urban (first column), road (second column) and highway (third column).}
\label{fig:comparisonFloat}
\end{figure}

Analyzing the numerical results of the reconstructed images, it can be seen that all the classes have a great IoU with the exception of the class Road Marks which suffers from a low precision value; in the first experiment, in particular, for every 100 TPs of its class there are 37 FPs. However, as Road Marks is the minority class, this value does not affect the overall result as Figure \ref{fig:comparisonFloat} shows.

Figure \ref{fig:comparisonFloat} also depicts how the proposed FCN perfectly segments a typical driving scene (second column) for the 3-class experiment (second and third rows) while it fails to correctly identify some pixels in challenging images such as those where there are objects casting their shadows on the road (first column, especially in the background) or overlapping objects (third column, where the left side is populated with objects of different materials).

Table \ref{tab:metricsFloatingPoint} also confirms the good segmentation of the 5-class experiment. The global result can be seen in Figure \ref{fig:comparisonFloat} where, once again, the overall segmentation would be very useful in a lane-keeping system and would also allow the driver to have more information about the surroundings. For instance, it can be observed that the system is now able to identify the presence of some objects in the no-drivable sections of the images such as traffic signals, pedestrians and guardrails.

\subsection{A Comparison with Baseline Spectral Classifiers}
From the above described results it can be concluded that the contribution of the spatial information provided by the convolution filters is, indeed, relevant to overcome the limitations inherent to the spectral separability of the different objects that can be present in real driving scenes. In order to get a more precise picture of this contribution we have compared the obtained results to those achieved with a baseline purely spectral classifier based on a three-hidden-layer feedforward ANN. The exploration and optimization process carried out has concluded with 25-25-100-100-3 being the best structure for the network. Table \ref{tab:metricsFloatingPoint} gathers the metrics for the two experiments of the three-hidden-layer neural network.

The results achieved by the ANN are, all in all, far from the ones associated to the U-Net (specially in terms of the precision of the minority class), so it can be confirmed that the joint use of spatial and spectral information is beneficial for the segmentation ability of a neural network. 

The difference in performance can also be explained in terms of model size and computational complexity (MACs, Multiply and Accumulate operations): while the ANN has only 13,855 parameters and performs 1,203,687,000 MACs (13,625 MACs per pixel) during inference, the U-Net has 31,725 parameters (320 of them are non-trainable) and needs 2,543,321,088 MACs (141,295,616 MACs per patch) to produce an output. The difference between the MACs ratio (2.11x) and the parameters ratio (2.3x) affects the time needed to make the forward pass, which will be assessed in the next section. 

It is worth mentioning a relevant outcome relative to the significance of the information provided by the spectral bands of the sensor. Despite the high correlation observed between spectral bands in the dataset images, we have verified that the reduction of the spectral bands used in the training of spectral classifiers strongly conditions the achievable accuracy of the segmentation.

\begin{figure}[h!]
\centering
\includegraphics[height=4.05cm]{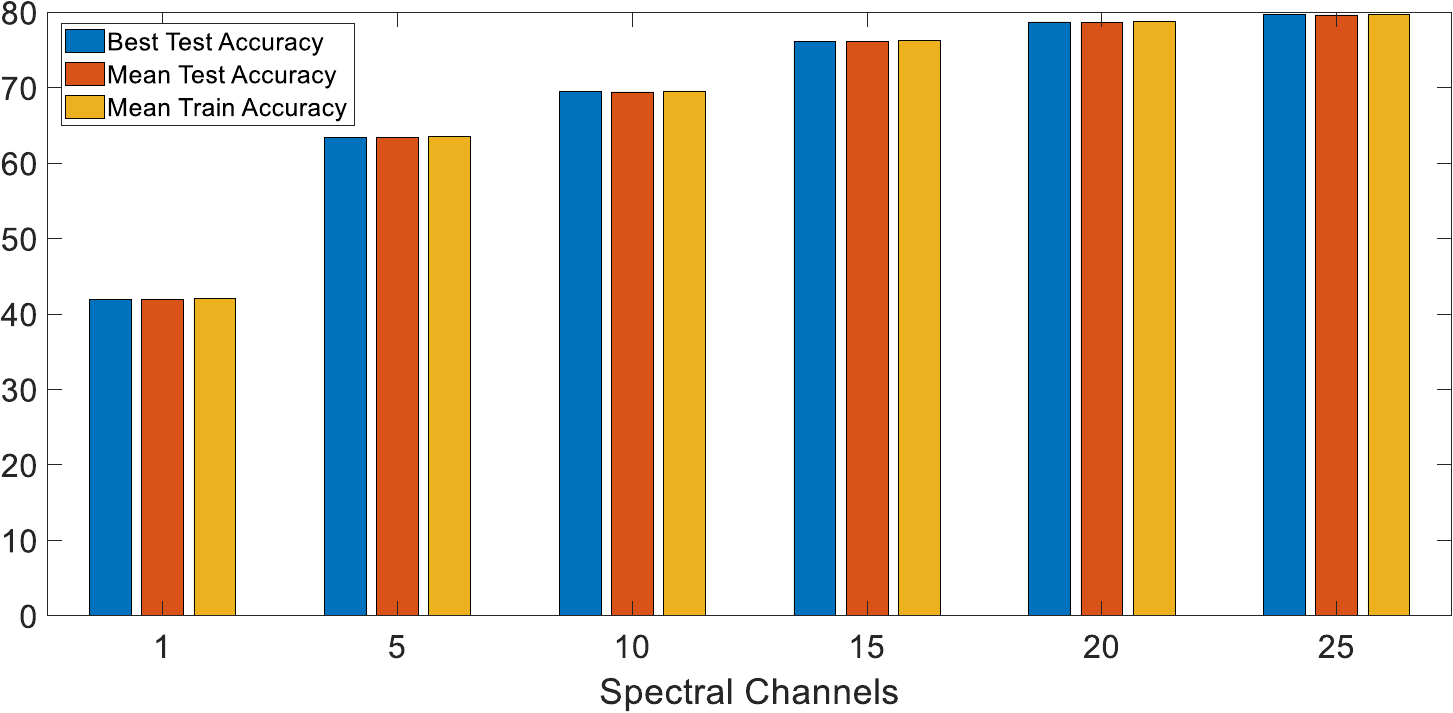}
\caption{Overall accuracy (\%) as a function of the number of spectral channels.}
\label{fig:barPlot}
\end{figure}

As an example, Figure \ref{fig:barPlot} shows the overall accuracy results of a reference ELM (Extreme Learning Machine) classifier for a different number of spectral bands (ranging from just 1 to all the 25 bands). It can be noted that accuracy indexes vary almost 40\%. On the contrary, we have observed that reducing the spectral bands in the training of the U-Net has not such a strong effect and the spatial information can compensate to a great extent the reduction in the spectral information provided as input. In particular, we have observed that when using just one spectral band (a gray level image, actually) the segmentation performance has only degraded, in terms of overall accuracy, by 0.75\% when that band is the first principal component of a PCA (Principal Component Analysis) and by 1.75\% when that band is just 1 of the 25 bands.

\section{Workflow for Rapid Prototyping}\label{sec:workflow}
As the implementation of this kind of neural networks in SoCs is a challenging and time consuming process, we have decided to explore the use of high-level automatic code generation tools to achieve a rapid prototyping of the system.

\subsection{Image Preprocessing}\label{subsec:imagePre}
Raw images acquired from mosaic snapshot cameras need to undergo a preprocessing pipeline in order to be converted into hyperspectral cubes. This process, which starts with raw image cropping and finishes with band normalization, needs to be taken into account when characterizing the throughput of the whole segmentation system. The rest of the steps are reflectance correction, partial demosaicing (original resolution is not restored), band alignment and spatial filtering. This processing has been codified in C language and compiled to be executed as an embedded Linux application in the microprocessor of the MPSoC as part of the HW/SW codesign for the implementation of the system.

\begin{table}[h!]
\centering
\caption{Mean execution time of the image-preprocessing Linux application.}
\label{tab:profilingOS}
\resizebox{5.5cm}{!}{%
\begin{tabular}{c|c|}
\hline
\multicolumn{1}{|c|}{\textbf{Step Name}} & \textbf{Execution time (ms)} \\ \hline
\multicolumn{1}{|c|}{Image cropping} & 4.36 \\ \hline 
\multicolumn{1}{|c|}{Reflectance correction} & 68.49 \\ \hline
\multicolumn{1}{|c|}{Partial demosaicing} & 30.15 \\ \hline
\multicolumn{1}{|c|}{Band alignment} & 22.41 \\ \hline
\multicolumn{1}{|c|}{Spatial filtering} & 202.65 \\ \hline
\multicolumn{1}{|c|}{Band normalization} & 26.06 \\ \hline
\multicolumn{1}{|c|}{Total} & 353.97 \\ \hline
\end{tabular}}
\end{table}

Table \ref{tab:profilingOS} shows the mean latency over 1000 iterations of the preprocessing pipeline running on the Cortex A-53 Quadcore processor (which has NEON SIMD extensions mandatory per core) in the Zynq MPSoC which reaches a reasonable value of 353.97 ms.

\subsection{Neural Network Deployment}
The design, training, validation and test of the FCN has been performed using MATLAB's Deep Network Designer. For the segmentation system prototyping process on the MPSoC we have used Vitis AI, a development platform for AI inference on Xilinx hardware platforms. Since there is not a direct procedure for the implementation of MATLAB-generated deep models we had to first export the neural network to open neural network exchange (ONNX) representation and then import it to Keras, our chosen framework, via \textit{onnx2keras}, an ONNX to Keras neural network converter \cite{onnx2keras}. The next steps are the freezing and quantizating processes of the neural network that are necessary due to the fact that Vitis AI favours integer computing.

The workflow continues by creating a Tensorflow inference graph from the Keras model and by removing the unnecessary information from training and saving only the required elements to compute the requested outputs. As it is known, inference is computationally expensive and, in order to reach the high-throughput and low-latency requirements of ADAS applications, a high memory bandwidth is required. Vitis AI Quantizer exploits quantization and VAI Optimizer applies channel pruning techniques to meet those issues.

According to \cite{quantModel}, by converting the 32-bit floating point weights and activations to 8-bit integer format, Vitis AI quantizer can reduce computing complexity without losing prediction accuracy and, as the fixed-point network model requires less memory bandwidth, a faster speed is provided. 

Finally, the product of the quantization is loaded at runtime in the system composed by the ARM CPU and the DPU accelerator in the MPSoC by a Python VART API. Although the quantization output of the 3-class experiment presents no appreciable IoU degradation, it has to be stated that the 5-class quantized model experiences a noticeable loss of performance on some images, an issue to be addressed in the future by Quantization Aware Training or Finetuning \cite{quantModel}.

\begin{figure}[h!]
\begin{subfigure}{0.32\linewidth}
\centering
\includegraphics[height=1.9cm]{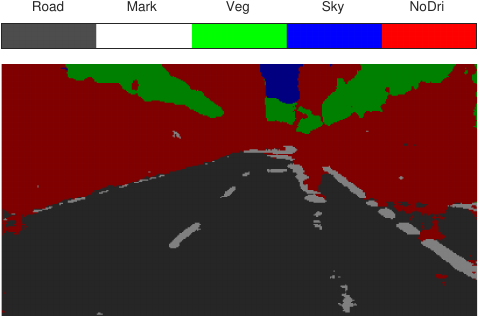}
\caption{Urban (OIoU, 93.78).}
\label{fig:segUrban}
\end{subfigure}
\begin{subfigure}{0.32\linewidth}
\centering
\includegraphics[height=1.9cm]{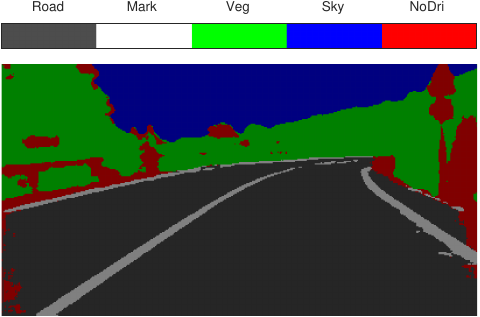}
\caption{Road (OIoU, 93.61).}
\label{fig:segRoad}
\end{subfigure}
\begin{subfigure}{0.32\linewidth}
\centering
\includegraphics[height=1.9cm]{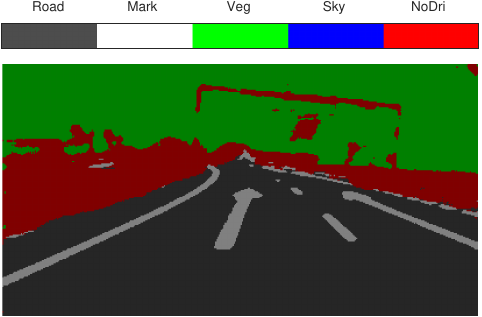}
\caption{Highway (OIoU, 94.96).}
\label{fig:segHighway}
\end{subfigure}
\caption{Segmented images produced by the deployed model on the MPSoC.}
\label{fig:deployedUnet}
\end{figure}

Figure \ref{fig:deployedUnet} shows the output of the deployed model and confirms its good overall performance. In fact, the similarity between quantized and unquantized results is 97.82\%, 98.16\% and 98.66\%, respectively.

In terms of throughput, the inference of the U-Net deployed in the Zynq UltraScale+ MPSoC reaches a rate of 487.91 FPS when 2 DPUCZDX8G cores (B4096 architecture) are involved, that is, an image (18 patches) is segmented every 36.89 ms (27 FPS). In \cite{courdier2020real} the FPS of some of the state-of-the-art neural networks for segmentation are evaluated when run on a GTX 1080 Ti and range from 5 to 23 FPS. However, if we add to it the time employed during the preprocessing (Table \ref{tab:profilingOS}), the throughput decreases to 2.55 FPS which can be considered a good starting point but will need to be improved by accelerating by hardware some of the preprocessing steps or changing the band extraction approach.

We have also benchmarked three different device types and evaluated their performance in terms of throughput with comparative purposes: an Intel-Xeon E5 1620 v3 Quadcore (CPU) and two embedded platforms such as a Jetson Nano (GPU) and a Xilinx ZCU104 (FPGA).

\begin{table}[h!]
\centering
\caption{Throughput of the U-Net and ANN on different device types}
\label{tab:throughput}
\resizebox{11cm}{!}{
\begin{tabular}{c|cccc|cccc|}
\cline{2-9}
 &
  \multicolumn{4}{c|}{\textbf{U-Net (18x128x128x25)}} &
  \multicolumn{4}{c|}{\textbf{ANN (216x409x25)}} \\ \hline
\multicolumn{1}{|c|}{\textbf{\diagbox[]{Device type}{Timing (FPS)}}} &
  \multicolumn{1}{c|}{\textbf{Mean}} &
  \multicolumn{1}{c|}{\textbf{Median}} &
  \multicolumn{1}{c|}{\textbf{Max}} &
  \textbf{Min} &
  \multicolumn{1}{c|}{\textbf{Mean}} &
  \multicolumn{1}{c|}{\textbf{Median}} &
  \multicolumn{1}{c|}{\textbf{Max}} &
  \textbf{Min} \\ \hline
\multicolumn{1}{|c|}{\textbf{CPU (Intel Xeon), FP32}} &
  \multicolumn{1}{c|}{1.49} &
  \multicolumn{1}{c|}{1.51} &
  \multicolumn{1}{c|}{1.62} &
                      0.19  &
  \multicolumn{1}{c|}{2.49} &
  \multicolumn{1}{c|}{2.51} &
  \multicolumn{1}{c|}{3.29} &
                      1.28  \\ \hline
\multicolumn{1}{|c|}{\textbf{GPU (Jetson Nano), FP16/FP32}} &
  \multicolumn{1}{c|}{10.70} &
  \multicolumn{1}{c|}{10.76} &
  \multicolumn{1}{c|}{11.80} &
                       3.07  &
  \multicolumn{1}{c|}{17.74} &
  \multicolumn{1}{c|}{17.90} &
  \multicolumn{1}{c|}{19.50} &
                       6.12  \\ \hline
\multicolumn{1}{|c|}{\textbf{FPGA (Zynq Ultrascale+), INT8}} &
  \multicolumn{1}{c|}{27.11} &
  \multicolumn{1}{c|}{27.70} &
  \multicolumn{1}{c|}{29.39} &
                      19.52  &
  \multicolumn{1}{c|}{-} &
  \multicolumn{1}{c|}{-} &
  \multicolumn{1}{c|}{-} &
                      -  \\ \hline
\end{tabular}}
\end{table}

Table \ref{tab:throughput} displays the mean, median, maximum and minimum FPS values of 1000 executions of the segmentation of one image with both neural networks on each device. It shows, on the one hand, how the ANN is 1.66x faster compared to the U-Net although we have to take into account that U-Net performs only 2.11x MACs to get a much better segmentation output. On the other hand, it also confirms the benefits of using a dedicated custom hardware such as the DPU included in the PL part of the MPSoC which outperforms the U-Net CPU and GPU approaches (18.2x and 2.5x FPS ratios respectively) and even exceeds the ANN GPU alternative (1.5x FPS ratio). 

\section{Conclusions}\label{sec:conc}
The incorporation of richer spectral information through HSI improves the segmentation results of purely spectral models. Besides, it is confirmed that the use of spatial information via convolution operations outperforms purely spectral models, even when dealing with images as intricate and heterogeneous as those that must be processed in real driving scenarios. However, the contribution of the spectral information in spectro-spatial convolutional models needs to be further investigated since our experiments reveal that the spectral information is being overshadowed by the spatial information in the training process of the FCN segmentation. We expect that the more effective incorporation of the spectral information to the AI models should improve segmentation performance in tricky situations such as when there are areas with shadows, there is degradation in the materials to be segmented, there are surfaces with very high reflectance in conditions of extreme lighting or there are multiple overlapping objects.

The improvement of the segmentation performance involves investigating further modifications to the proposed U-Net such as using 3D convolutions or applying multiscale convolution techniques to extract spatial features at different scales. The use of different image preprocessing techniques (modifying the partial demosaicing step, for example) and the addition of a postprocessing stage (not to label pixels with uncertain prediction, for instance) will also be assessed, but their applicability will always be subject to the demanding throughput requirements of ADAS/ADS.

In turn, a workflow which combines the use of MATLAB's (Deep Learning and Image Processing toolboxes) and Xilinx's tools (AI development and deployment environments) for the rapid prototyping of AI applications has been set. This will allow us to perform agile deployment characterization of future models so as to rapidly evaluate their suitability to ADAS applications. 

Finally, we have verified that the combination of an encoder-decoder FCN and the prototyping platform (Zynq Ultrascale MPSoC) allows us to perform the preprocessing of the cubes and the segmentation of the images in a reasonable time for this kind of applications outperforming CPU-only and GPU approaches. However, there are multiple pathways for future research, either in terms of the acceleration of the cube generation (hardware acceleration of the median filtering) or the optimization of the U-Net. At the same time, a deeper investigation is required regarding the data transfer between memory and the different components of the SoC to try to improve the throughput and reduce the cost and consumption.

\bibliographystyle{splncs04}
\bibliography{biblio.bib}

\end{document}